# A Novel Approach to Skew-Detection and Correction of English Alphabets for OCR

Skew-Detection using COG method and Correction using Sub-Pixel Shifting method

Chinmay Chinara[1]
Department of ECE
ITER, SOA University
Bhubaneswar, Odisha, India
chinmay_chinara@yahoo.co.in

Nishant Nath[2]
Department of ECE
ITER, SOA University
Bhubaneswar, Odisha, India
nishant.nath91@gmail.com

Subhajeet Mishra[3]
Department of EIE
ITER, SOA University
Bhubaneswar, Odisha, India
subhajeet.mishra7@gmail.com

Sangram K. Sahoo[4]
Department of EE
ITER, SOA University
Bhubaneswar, Odisha, India
shockwavesangram@gmail.com

Farida A. Ali[5]
Department of EIE
ITER, SOA University
Bhubaneswar, Odisha, India
farida.ashraf.ali@gmail.com

*Abstract*— Optical Character Recognition has been a challenging field in the advent of digital computers. It is needed where information is to be readable both to humans and machines. The process of OCR is composed of a set of pre and post processing steps that decide the level of accuracy of recognition. This paper deals with one of the pre-processing steps involved in the OCR process i.e. Skew (Slant) Detection and Correction. The proposed algorithm implemented for skew-detection is termed as the COG (Centre of Gravity) method and for that of skew-correction is Sub-Pixel Shifting method. The algorithm has been kept simple and optimized for efficient skew-detection and correction. The performance analysis of the algorithm after testing has been aptly demonstrated.

## I. INTRODUCTION

The field of Character Recognition is one of the major subsets of Pattern Recognition. Over the years, offline character recognisation has been achieving great demand due to evolution in the field of digital library and banking. The major advantage of these off-line recognizers is to allow the previously written and printed texts to be processed and recognized. The recognition accuracy totally depends on the quality of the processed characters. The more constrained the characters are the better will be the recognition.

Skew in an acquired image generally occurs due to the inaccuracies in the process of image acquisition. Skew-detection and correction is one of the most important pre-processing steps involved in the process of OCR. We will be dealing with the skew-detection and correction of textual images i.e. images containing textual information. At present, it is required to be more robust to deal with target images like camera-captured textual images in addition to regular business document images by scanner. Generally, there are two basic types of textual images over which skew-detection is applied. One type refers to those which have the whole text in the image gone topsy-turvy as shown in Figure 1. The other type refers to the one in which the textual content is straight but the written text is in the form of italics as shown in Figure 2. This paper deals with the latter kind of textual image in which the skew angle of each alphabet present in the text is studied individually and finally the necessary correction is made as per the algorithm suggested.

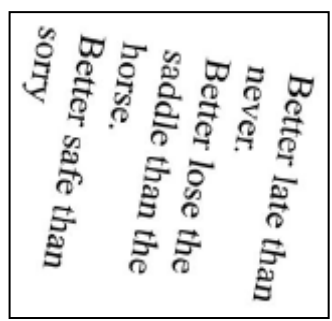
Figure 1. Topsy-turvy textual image

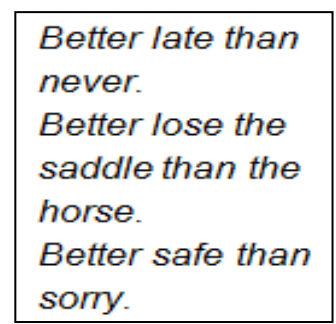
Figure 2. Italic textual image

This paper is organized as follows: Section-II gives the background about the Modern English and its diversity along with a brief idea on typeface. Section-III gives the basic literature review over the different methodologies used in the process of slant-estimation till date. In Section-IV, the proposed algorithm is described in detail, while Section-V presents the results and conclusion. Finally, Section-VI deals with the development platform used for the testing of algorithms.

## II. BACKGROUND – THE MODERN ENGLISH

The modern English alphabets are a sub-group of the 26 Latin alphabets set by the International Organization of Standardization (ISO). The shape of these alphabets depends on the typeface, commonly known as the font of the alphabet. The above 26 alphabets are represented in two forms, one being the Majuscule form (also called Uppercase or Capital letters) and the other being the Minuscule form (also called Lowercase or Small letters). These are again sub-divided into two parts, namely Vowels and Consonants. The characters consist of a vast database and they vary from typeface to typeface. The few alphabets and characters over which skew-detection and correction was implemented are shown in Figure 3 and Figure 4.They are based on the Arial font.

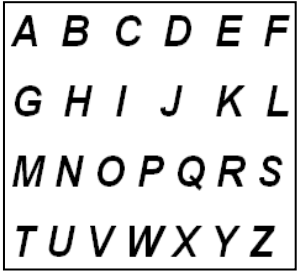 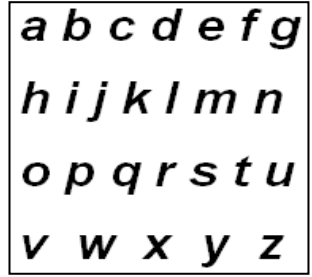

Figure 3. Majuscule form of alphabet (italics)     Figure 4. Minuscule form of alphabet (italics)

## III. LITERATURE SURVEY

There are several conventional methods for skew estimation and correction. In this section, we discuss some of the methods that have been implemented till date.

In [1], the slant angle is estimated by the inclination of the line connecting the gravity centers of the top 25% part and the bottom 25% part of the image. Then a sub-pixel shear transformation is performed in order to remove the estimated inclination.

In [2], the Hough Transform method is used for the process of slant angle estimation and thus correcting the slant.

In [3], a part-based skew estimation method has been implemented which is more robust to larger varieties of text images. Here, the skew angle at each local part of the input image is estimated independently by referring the local part of upright character images stored as a database. The global skew angle is then estimated by aggregating the estimated local skews. Its robustness lies in the fact that this method does not assume that the characters are laid out in straight lines.

In [4], the method proposed is a gradient based on combination with a Focused Nearest Neighbor Clustering (FNNC) of interest points that has no limitations regarding the detectable angle range. The upside-down decision is based on statistical analysis of ascenders and descenders. It can be applied to the entire document as well as its fragments.

In [5], the slant estimation is done based on analyzing the non-horizontal parts of the characters. The orientation, height of the bounding-box that includes all non-horizontal parts as well as their location related to the word. The non-horizontal parts orientation is estimated by weighing according to the height of the corresponding bounding box. An additional weight is applied if the fragment is outside of the core-region of the word, which indicates that this fragment is probably one of the strokes that should be, by definition, vertical to text orientation.

## IV. THE PROPOSED ALGORITHM

The skew-detection was implemented based upon the algorithm proposed by [1] in which the skew angle was estimated as the inclination of the line connecting the gravity centers of the top 25% part and the bottom 25% part of the acquired textual image. The textual image was first scanned vertically for obtaining the upper 25% and the lower 25%. The skew basically occurred in the foreground pixels (black). So, the basic aim was to find the center of gravity of the black pixels present each of the upper and lower portions of the textual image.

The skew-correction techniques proposed by [1] use the method of sub-pixel precision shear transformation for processing of textual images. The proposed algorithm for skew-correction is based on a new yet simple approach i.e. sub-pixel shifting.

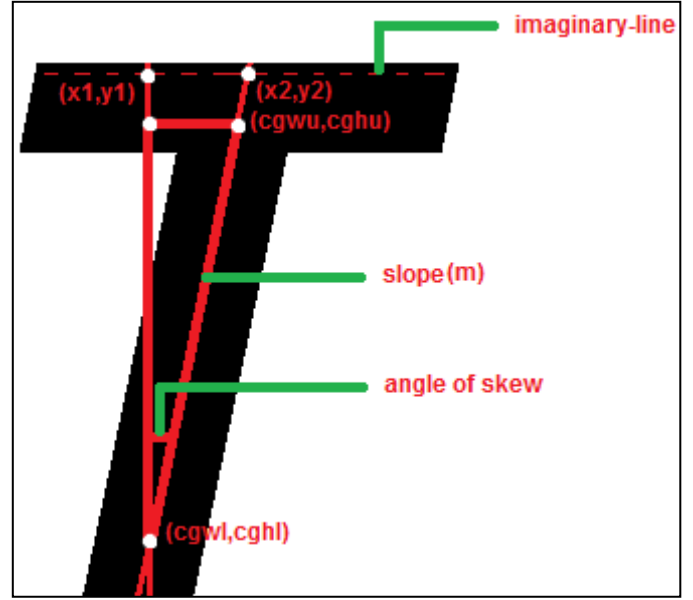

Figure 5. Reference for all the calculations done in the proposed algorithm

### A. Calculation of center of gravity of the upper 25%:

Considering the upper 25%, the image was then scanned horizontally in two halves, one half ranging from extreme left to the centre (left zone) and the other half ranging from the centre to the extreme right (right zone). In each of these zones, the number of black pixels present was calculated. Also, the maximum value of black-pixel location (represented by *maxwu*) and minimum value of black-pixel location (represented by *minwu*) was found by horizontal scanning. The width co-ordinate component of the center of gravity of the upper 25% (represented by *cgwu*) of the character image was calculated based on the following conditions:

i. if (black_pixel [left-zone] > black_pixel [right-zone])
      cgwu=minwu+((maxwu-minwu)/4)
ii. if (black_pixel [left-zone] < black_pixel [right-zone])
      cgwu=minwu+((maxwu-minwu)/2)
iii. if (black_pixel [left-zone] = = black_pixel [right-zone])
      cgwu=minwu+((maxwu-minwu)/2);

The height co-ordinate component of the center of gravity of the upper 25% (represented by *cghu*) was calculated as *(upper 25% image height)/2*.

### B. Calculation of center of gravity of the lower 25%:

For the lower 25%, the calculation of black-pixels in each zone and calculation for the maximum (represented by

*maxwl*) and minimum (represented by *minwl*) values of black-pixel location by horizontal scanning was done in the same process as done for the upper 25%. The co-ordinate of the center of gravity of the lower 25% was calculated as *(lower 25% image height)/2* (for height represented by *cghl*) and *minwl+((maxwl-minwl)/2* (for width represented by *cgwl*). The centers of gravity of the upper 25% and lower 25% were then joined by a line.

*C. Determination of the angle of skew:*

The character images taken have a maximum skew in the upper half rather than the lower half. So, the angle of skew is determined considering the upper portion as the base and taking the right or left side of the image as perpendicular depending upon the direction of skew. The skew-angle was calculated based on the following formula:

$$Skew - angle = \frac{180 * \tan^{-1}\left(\frac{length\ of\ perpendicular}{length\ of\ base}\right)}{\pi}$$

The skew-angle detection for some of the Majuscule alphabets has been shown in figures given below:

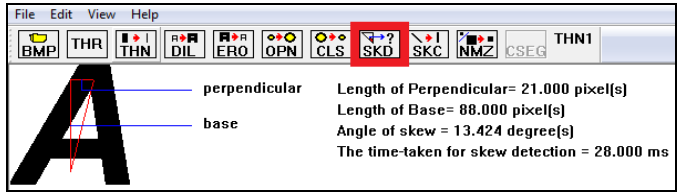

Figure 6. Skew-detection of the italic alphabet 'A'

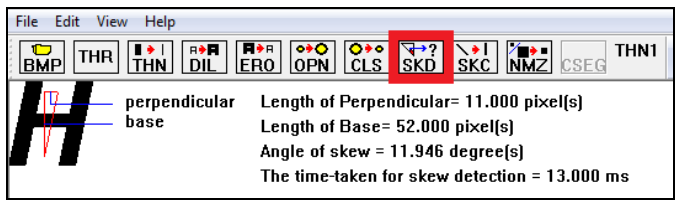

Figure 7. Skew-detection of the italic alphabet 'H'

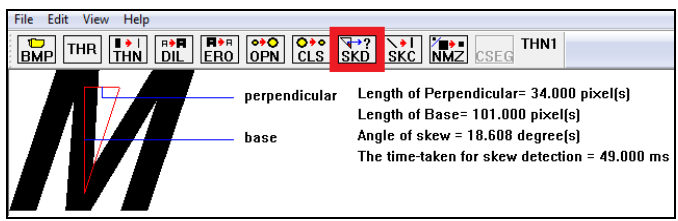

Figure 8. Skew-detection of the italic alphabet 'M'

*D. Determination of the co-ordinates of all the pixels lying on the line connecting the center of gravity and skew-correction*:

The line connecting the centers of gravity *(cgwl, cghl)* and *(cgwu, cghu)* was extended on both sides till it touches the top and bottom rows. Then a line passing straight through the center of gravity of the lower 25% was made which acted as the reference normal. The character image was then scanned horizontally considering each row as a line. The co-ordinates of this imaginary line intersected the normal-line at *(x1, y1)* and the extended COG line at *(x2, y2)*. '*x1*' had a value of '*cgwl*', which represented the normal-line. '*y1*' and '*y2*' varied from 0 to the height of the image. For calculating '*x2*', the slope of the COG line has to be calculated (denoted by '*m*'):

$$m = \left|\frac{cghu - cghl}{cgwu - cgwl}\right|$$

$$x2 = \frac{cghu + (height\ of\ image) - cghl + (m * cgwl)}{m}$$

The distance between the points '*x2*' and the point '*cgwl*' was recorded for each value of '*y2*'. The sub-pixel shifting was then done row by row starting from top to bottom, based on the distances obtained. If the character image tilted towards the left, sub-pixel shifting was done to the right and if it tilted to the right then shifting was done to the left.

The skew-angle correction for some of the Majuscule alphabets has been shown in figures given below:

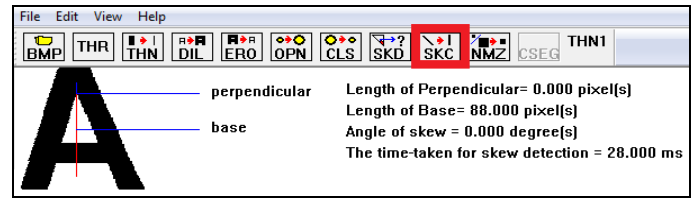

Figure 9. Skew-corrected italic alphabet 'A'

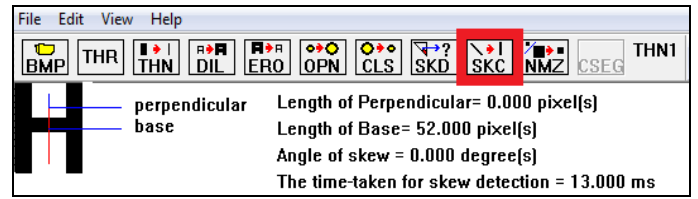

Figure 10. Skew-corrected italic alphabet 'H'

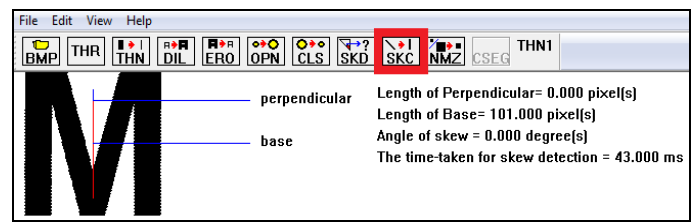

Figure 11. Skew-corrected italic alphabet 'M'

From the above figures it can be seen that the skew angle is zero which means that the alphabet has been perfectly skew-corrected.

## V. RESULTS AND CONCLUSION

The algorithm was tested with all the Majuscule and Minuscule form of alphabets with Arial font size 72. The size constraint was taken into account so as to make the results clearly visible on the screen. This algorithm however works very well for any size of alphabet taken. The table below shows the slant-angle measurement before and after slant correction.

TABLE 1. ACCURACY IN SLANT-CORRECTION FOR THE MAJUSCLE SET OF ALPHABETS

| Majuscule Alphabets (in italics) | Slant angle before skew-correction (in degrees) | Slant Angle after skew-correction (in degrees) | Accuracy (in %) | Average time-taken (in ms) |
|---|---|---|---|---|
| A | 13.424 | 0 | 100 | 28 |
| B | 10.454 | 0 | 100 | 27 |
| C | 11.378 | 0 | 100 | 29 |
| D | 14.399 | 0 | 100 | 32 |
| E | 12.638 | 0 | 100 | 37 |
| F | 11.347 | 2.303 | 67.32 | 18 |
| G | 13.439 | 0 | 100 | 24 |
| H | 11.946 | 0 | 100 | 13 |
| I | 12.958 | 0 | 100 | 12 |
| J | 10.112 | 1.245 | 87.68 | 23 |
| K | 9.821 | 0 | 100 | 34 |
| L | 14.543 | 0.901 | 93.80 | 32 |
| M | 18.608 | 0 | 100 | 46 |
| N | 19.000 | 0 | 100 | 20 |
| O | 12.343 | 0 | 100 | 24 |
| P | 14.424 | 1.345 | 90.675 | 25 |
| Q | 12.346 | 0 | 100 | 43 |
| R | 13.816 | 0 | 100 | 28 |
| S | 12.116 | 0 | 100 | 29 |
| T | 17.829 | 0 | 100 | 33 |
| U | 10.989 | 0 | 100 | 34 |
| V | 10.978 | 0 | 100 | 36 |
| W | 18.699 | 0 | 100 | 21 |
| X | 19.787 | 0 | 100 | 35 |
| Y | 13.700 | 0 | 100 | 22 |
| Z | 12.638 | 0 | 100 | 27 |

As per the results shown in the table, the algorithm worked perfectly for all the alphabets except for F, J, L and P. This occurred mainly due to the uneven distribution of pixels on either side of the COG of the alphabet, which is the base of the skew-detection algorithm proposed. However, the time-complexity of the process is very optimum (in the range of milliseconds). So, this algorithm can be very well be used for skew detection and correction, even though there is room for improvement.

## VI. DEVELOPMENT PLATFORM

The algorithm was implemented using the Microsoft Visual C++ (ver. 2010) development platform. Due to its quick, enhanced and interactive environment, we were able to study, implement and analyze the different steps involved in the proposed algorithm. The processor used was *Intel Core 2 Duo, 2.20 GHz, 4.00 GB RAM, 512MB ATI Radeon Graphics*.

## ACKNOWLEDGEMENT

We acknowledge the research facilities and development platform provided to us at the Video-Data Analysis Systems Lab, Electro-Optical Tracking Division, Integrated Test Range, DRDO, Chandipur for the successful completion of our research.

We would also like to thank our faculty and project-in-charge Ms. Farida Ashraf Ali, Asst. Professor, Dept. of EIE, ITER, SOA University for her continuous guidance and support over the period of research.